\documentclass[11pt,a4paper]{article}
\usepackage[hyperref]{emnlp2020}
\usepackage{times}
\usepackage{latexsym}
\usepackage{graphicx}
\usepackage{xspace}
\usepackage{titlecaps}
\usepackage{multirow}
\usepackage{comment}
\usepackage{CJKutf8}

\newcommand{\xlwic}{XL-WiC\xspace}
\newcommand{\lang}[1]{\uppercase{#1}\xspace}
\newcommand{\bertde}{BERT-de\xspace}
\newcommand{\bertit}{BERT-it\xspace}
\newcommand{\bertfr}{CamemBERT-large\xspace}
\newcommand{\lbert}{L-BERT\xspace}
\newcommand{\en}{\lang{en}}
\newcommand{\bg}{\lang{bg}}
\newcommand{\hr}{\lang{hr}}
\newcommand{\da}{\lang{da}}
\newcommand{\de}{\lang{de}}
\newcommand{\et}{\lang{et}}
\newcommand{\fa}{\lang{fa}}
\newcommand{\fr}{\lang{fr}}
\newcommand{\itlang}{\lang{it}}
\newcommand{\ja}{\lang{ja}}
\newcommand{\ko}{\lang{ko}}
\newcommand{\nl}{\lang{nl}}
\newcommand{\zh}{\lang{zh}}
\usepackage{subcaption}

\newcommand{\mbert}{mBERT\xspace}
\newcommand{\xlmrlarge}{XLMR-large\xspace}
\newcommand{\xlmrbase}{XLMR-base\xspace}

\newcommand{\alllangs}{Bulgarian, Chinese, Croatian, Danish, Dutch, Estonian, Farsi, French, German, Italian, Japanese and Korean\xspace}
\newcommand{\alllangsAndCode}{Bulgarian (BG), Chinese (ZH), Croatian (HR), Danish (DA), Dutch (NL), Estonian (ET), Farsi (FA), French (FR), German (DE), Italian (IT), Japanese (JA) and Korean (KO)\xspace}

\newcommand{\zeroshot}{zero-shot\xspace}
\usepackage{microtype}
\usepackage{booktabs}
\usepackage{siunitx}

\usepackage{setspace}

\newcommand\blfootnote[1]{%
  \begingroup
  \renewcommand\thefootnote{}\footnote{#1}%
  \addtocounter{footnote}{-1}%
  \endgroup
}

\usepackage{paralist}
\aclfinalcopy % Uncomment this line for the final submission

\title{XL-WiC: A Multilingual Benchmark \\for Evaluating Semantic Contextualization}

\author{
Alessandro Raganato$^{\star\heartsuit}$ ~ Tommaso Pasini$^{\star\diamondsuit}$  \\
\bf{Jose Camacho-Collados}$^{\clubsuit}$ ~ \bf{Mohammad Taher Pilehvar}$^{\spadesuit}$ \\
$^\heartsuit$ Department of Digital Humanities, University of Helsinki, Finland \\
$^\diamondsuit$ SapienzaNLP Group, Computer Science Department, Sapienza University of Rome, Italy \\
$^\clubsuit$ School of Computer Science and Informatics, Cardiff University, United Kingdom \\
$^\spadesuit$ Tehran Institute for Advanced Studies, Iran \\
{\tt $^{\heartsuit}$alessandro.raganato@helsinki.fi}, 
{\tt $^{\diamondsuit}$pasini@di.uniroma1.it},\\ 
{\tt $^{\clubsuit}$camachocolladosj@cardiff.ac.uk}, 
{\tt $^{\spadesuit}$mp792@cam.ac.uk}
}

\date{}

\begin{document}
\maketitle
\begin{abstract}
The ability to correctly model distinct meanings of a word is crucial for the effectiveness of semantic representation techniques.
However, most existing evaluation benchmarks for assessing this criterion are tied to sense inventories (usually WordNet), restricting their usage to a small subset of knowledge-based representation techniques.
The Word-in-Context dataset (WiC) addresses the dependence on sense inventories by reformulating the standard disambiguation task as a binary classification problem; but, it is limited to the English language.
We put forward a large multilingual benchmark, XL-WiC, featuring gold standards in 12 new languages from varied language families and with different degrees of resource availability, opening room for evaluation scenarios such as zero-shot cross-lingual transfer.
We perform a series of experiments to determine the reliability of the datasets and to set performance baselines for several recent contextualized multilingual models.
Experimental results show that even when no tagged instances are available for a target language, models trained solely on the English data can attain competitive performance in the task of distinguishing different meanings of a word, even for distant languages. 
XL-WiC is available at \url{https://pilehvar.github.io/xlwic/}.
%Also, XLMR significantly outperforms multilingual BERT.
\blfootnote{Authors marked with a star ($^\star$) contributed equally.}

\end{abstract}

%\begin{spacing}{0.99}

\section{Introduction}

One of the desirable properties of contextualized models, such as BERT \cite{BERT} and its derivatives, lies in their ability to associate dynamic representations to words, i.e., embeddings that can change depending on the context.
This provides the basis for the model to distinguish different meanings (senses) of words without the need to resort to an explicit sense disambiguation step.
The conventional evaluation framework for this property has been Word Sense Disambiguation \cite[WSD]{navigli:09}.
%However, given that WSD is usually defined against a sense inventory, often WordNet \cite{}, the usage of the benchmark is limited to a subset of techniques that explicitly model word senses in the inventory, leaving out almost all the recent language model-based representation models.
%
%\newcite{pilehvar2019-wic} proposed a dataset, called WiC, that addresses this issue by reformulating WSD as an inventory-independent binary classification task. -SuperGLUE, etc.
%
However, evaluation benchmarks for WSD are usually tied to  external sense inventories (often WordNet \cite{Fellbaum:98}), making it extremely difficult to evaluate systems that do not explicitly model sense distinctions in the inventory, effectively restricting the benchmark to inventory-based sense representation techniques and WSD systems.
This prevents a direct evaluation of lexical semantic capacity for a wide range of inventory-free models, such as the dominating language model-based contextualized representations.

\newcite{pilehvar2019-wic} addressed this dependence on sense inventories by reformulating the WSD task as a simple binary classification problem: given a target word $w$ in two different contexts, $c_1$ and $c_2$, the task is to identify if the same meaning (sense) of $w$ was intended in both $c_1$ and $c_2$, or not.
The task was framed as a dataset, called Word-in-Context (WiC), which is also a part of the widely-used SuperGLUE benchmark \cite{wang2019superglue}.  
Despite allowing a significantly wider range of models for direct WSD evaluations, WiC is limited to the English language only, preventing the evaluation of models in other languages and in cross-lingual settings.

%In this paper we put forward a new evaluation benchmark....
In this paper, we present a new evaluation benchmark, called XL-WiC
%\footnote{XL-WiC is available at \red{\url{https://pilehvar.github.io/xlwic/}}.}
, that extends the WiC dataset to 12 new languages from different families and with different degrees of resource availability: \alllangsAndCode.
%enabling the evaluation of multilingual models in this challenging disambiguation task. %, introducing testbeds for other interesting scenarios}
With over 80K instances, our benchmark can serve as a reliable evaluation framework for contextualized models in a wide range of heterogeneous languages.
XL-WiC can also serve as a suitable testbed for cross-lingual experimentation in settings such as zero-shot or few-shot transfer across languages.
%{\color{blue}Two sentences on main conclusions from the experiments.}
As an additional contribution, we tested several pretrained multilingual models on XL-WiC, showing that they are generally effective in transferring sense distinction knowledge from English to other languages in the zero-shot setting. 
However, with more training data at hand for target languages, monolingual approaches gain ground, outperforming their multilingual counterparts by a large margin.

\section{Related Work}

XL-WiC is a benchmark for inventory-independent evaluation of WSD models (Section \ref{rel:WSD}), while the multilingual nature of the dataset makes it an interesting resource for experimenting with cross-lingual transfer (Section \ref{rel:cross}).

\subsection{Word Sense Disambiguation}
\label{rel:WSD}

The ability to identify the intended sense of a polysemous word in a given context is one of the fundamental problems in lexical semantics. It is usually addressed with two different kinds of approaches relying on either sense-annotated corpora \cite{bevilacquaandnavigli-2020-ewiser,scarlinietal:20,terraluke-2020-bem} or knowledge bases \cite{Moroetal:14tacl,agirreetal:14,scozzafava-etal-2020-syntagrank}. Both are usually evaluated on dedicated benchmarks, including at least five WSD tasks in Senseval and SemEval series, from 2001 \cite{edmonds-cotton-2001-senseval} to 2015 \cite{moro-navigli-2015-semeval} that are included in the \citet{raganatoetal:17}'s test suite.
%, merged by  \newcite{raganato-camachocollados-navigli:2017:EACLlong} into a unified framework, currently used for WSD.
All these tasks are framed as classification problems, where disambiguation of a word is defined as selecting one of the predefined senses of the word listed by a sense inventory.
This brings about different limitations such as restricting senses only to those defined by the inventory, 
%or making the WSD system dependent on the inventory,
%as it needs to explicitly model sense distinctions in the inventory in order to be able to pick one
or forcing the WSD system to explicitly model sense distinctions at the granularity level defined by the inventory.% in order to be able to pick one.

%Word Sense Induction, ... adds the difficulty of mapping induces senses to the external sense inventory.

Stanford Contextual Word Similarity \cite{huang-etal-2012-improving} is one of the first datasets that focuses on ambiguity but outside the boundaries of sense inventories, and as a similarity measurement between two words in their contexts.
\newcite{pilehvar2019-wic} highlighted some of the limitations of the dataset that prevent a reliable evaluation, and proposed the Word-in-Context (WiC) dataset.
WiC is the closest dataset to ours, which provides around 10K instances (1400 instances for 1184 unique target nouns and verbs in the test set), but for the English language only.

\subsection{Cross-lingual NLP}
\label{rel:cross}
%{\color{red}Better to focus mainly on the (1) available multilingual benchmarks here, such as those semeval tasks, xtreme, word similarity, sts, ... and then finally very briefly say that (2) they are now required more than ever as pre-trained models have opened up new opportunities for cross lingual transfer. This currently mainly focuses on (2) which is not central to our benchmark paper.}

A prerequisite for research on a language is the availability of relevant evaluation benchmarks. %for that language.
Given its importance, construction of multilingual datasets has always been considered as a key contribution in NLP research and numerous benchmarks exist for a wide range of tasks, such as semantic parsing \cite{hershcovich-etal-2019-semeval}, word similarity
\cite{camacho-collados-etal-2017-semeval,barzegar-etal-2018-semr}, sentence similarity \cite{cer-etal-2017-semeval}, %\red{language inference \cite{conneau2018xnli}} 
or WSD \cite{navigli-etal-2013-semeval,moro2015semeval}.
A more recent example is XTREME \cite{hu2020xtreme}, a benchmark that covers around 40 languages in nine syntactic and semantic tasks.

%Research in NLP has traditionally been centered on the English language.
%One reason behind this is the lack of benchmarks in those languages, which prevents effective training and reliable validation of models. 
%Because of this, a new branch has emerged in recent years, which is cross-lingual NLP. 
%The main idea is that knowledge obtained from one language (mostly English) can be used in other less-resourced languages. 
%For this cross-lingual embeddings have been proved effective methods to align two or more languages \cite{ruder2019survey}, even with minimal supervision \cite{zhang-etal-2017-earth,conneau2018word,artetxeacl2018unsupervised}. 

On the other hand, pre-trained language models have recently proven very effective in transferring knowledge in cross-lingual NLP tasks \cite{BERT,XLMR}. 
This has further magnified the requirement for rigorous multilingual benchmarks that can be used as basis for this direction of research \cite{artetxe2020call}.
%To facilitate their integration and evaluation into different tasks, different efforts to create multilingual benchmarks of standard NLP tasks have been put forward \cite{conneau2018xnli,hu2020xtreme}.

%Need for data in languages other than English, cross-lingual NLP, talk about XGLUE and XTREME...

%\subsection{Zero-shot Learning}
\begin{table*}[ht]
    \centering
    \resizebox{\linewidth}{!}{
    \begin{tabular}{l lll l}
    \toprule
    \bf Lang. & 
    \bf ~~Target Word & \multicolumn{1}{l}{\bf ~~Sentence 1} & \multicolumn{1}{l}{\bf ~~Sentence 2} & \bf Label\\
  %  \midrule
 %   EN &
 %   ~~Land&
 %   ~~The pilot managed to land the airplane safely. &
 %   ~~The enemy landed several of our aircrafts. & False\\
     \midrule
    EN &
    ~~Beat&
    ~~We beat the competition.  &
    ~~Agassi beat Becker in the tennis championship. & True\\
    \midrule
    DA & 
    ~~Tro&
    ~~Jeg tror på det, min mor fortalte.&
    ~~Maria troede ikke sine egne øjne.&
    True
    \\
    ET &
    ~~Ruum &
    ~~Ühel hetkel olin väljaspool aega ja ruumi.&
    ~~Ümberringi oli lõputu tühi ruum. &
    True \\
    %FR & & 
    %~~Rangée&
    %~~L'homme assis à la troisième rangée. &
    %~~Une rangée d’arbres, de maisons, de voitures. &
    %False \\
    FR &
    ~~Causticité&
    ~~Sa causticité lui a fait bien des ennemis. &
    ~~La causticité des acides. &
    False \\
    KO & 
    \begin{CJK}{UTF8}{}
    \CJKfamily{mj}
    틀림
    \end{CJK} & 
    \begin{CJK}{UTF8}{}
    \CJKfamily{mj}
    틀림이 있는지 없는지 세어 보시오. \end{CJK} &
    \begin{CJK}{UTF8}{}
    \CJKfamily{mj}
     그 아이 하는 짓에 틀림이 있다면 모두 이 어미 죄이지요.
    \end{CJK} &
    False\\
    ZH & 
    \begin{CJK}{UTF8}{}
    \CJKfamily{mj}
    發
    \end{CJK} & 
    \begin{CJK}{UTF8}{}
    \CJKfamily{mj}
    建築師希望發大火燒掉城市的三分之一。
    \end{CJK} &
    \begin{CJK}{UTF8}{}
    \CJKfamily{mj}
    如果南美洲氣壓偏低，則印度可能發乾旱 
    \end{CJK} &
    True\\
    \raisebox{0.6\height}{FA} &
    \raisebox{0.3\height}{~~\includegraphics[scale=0.4]{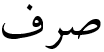}} &
    \includegraphics[scale=0.4]{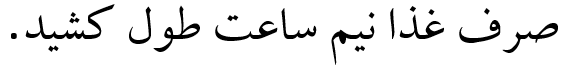} &
    \includegraphics[scale=0.4]{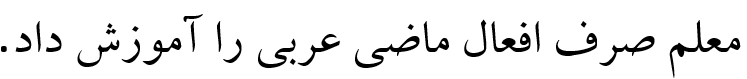} &
    \raisebox{.6\height}{False}
    \\
    \bottomrule
    \end{tabular}
    }
    \caption{Sample instances from XL-WiC for different languages. %\red{TODO: Would be nice to add one example per each language in the dataset (ideally verifying with a speaker of the language)}
    }
    \label{tab:xlwic_examples}
\end{table*}

\section{XL-WiC: The Benchmark}

In this section, we describe the procedure we followed to construct the \xlwic benchmark. Our framework is based on the original WiC dataset, which we extend to multiple languages.%inheriting many of its properties.
%such as its coarse-grained distinction of senses. (is it ok to add?)

\subsection{English WiC}

Each instance of the original WiC dataset \cite{pilehvar2019-wic} is composed of a target word (e.g., \textit{justify}) and two sentences where the target word occurs (e.g., ``{\it Justify} the margins'' and ``The end {\it justifies} the means''). 
The task is a binary classification: to decide whether the same sense of the target word (\textit{justify}) was intended in the two contexts or not. 
The dataset was built using example sentences from resources such as Wiktionary, WordNet \cite{miller1995wordnet} and VerbNet \cite{VerbNet}.

\subsection{XL-WiC}
We followed \newcite{pilehvar2019-wic} and constructed XL-WiC based on example usages of words in sense inventories.
Example usages are curated in a way to be self contained and clearly distinguishable across different senses of a word; hence, they provide a reliable basis for the binary classification task.
Specifically, for a word $w$ and for all its senses $\{s_1^w,...,s_n^w\}$, we extract from the inventory all the example usages.
We then pair those examples that correspond to the same sense $s_i^w$ to form a positive instance (True label) while examples from different senses (i.e., $s_i^w$ and $s_j^w$ where $i \ne j$) are paired as a negative instance (False label). 

%One of the main contributions of this paper is the extension of the WiC benchmark to a several other languages from different families and with different degrees of resource availability. 
We leveraged two main sense inventories for this extension: Multilingual WordNet (Section \ref{multilingual-wordnet}) and Wiktionary (Section \ref{wiktionary}).

\begin{table*}[!ht]
    \centering
    \setlength{\tabcolsep}{5.0pt}
    \resizebox{\linewidth}{!}{
    \begin{tabular}{llrrrrrrrrrrrrrrr}
    \toprule
    \multirow{2}{*}{\bf Split} & 
    \multirow{2}{*}{\bf Stat} &
     
    \multicolumn{1}{c}{\bf WiC} & \multicolumn{9}{c}{\bf Multilingual WordNet} &
    & 
    
    \multicolumn{3}{c}{\bf Wiktionary} \\
    %\cmidrule(lr){3-12} %\cmidrule(lr){13-16}
    \cmidrule(lr){3-3}
    \cmidrule(lr){4-13}
    \cmidrule(lr){14-16}
    
    &
    &
    \multicolumn{1}{c}{\bf \en} & 
    \multicolumn{1}{c}{\bf \bg} & 
    \multicolumn{1}{c}{\bf \da} & 
    \multicolumn{1}{c}{\bf \et} & 
    \multicolumn{1}{c}{\bf \fa} & 
    \multicolumn{1}{c}{\bf \hr} & 
    %\multicolumn{1}{c}{\bf \itlang} & 
    \multicolumn{1}{c}{\bf \ja} & 
    \multicolumn{1}{c}{\bf \ko} & 
    \multicolumn{1}{c}{\bf \nl} & 
    \multicolumn{1}{c}{\bf \zh} &
    & 
    \multicolumn{1}{c}{\bf \de} & 
    \multicolumn{1}{c}{\bf \fr} & 
    \multicolumn{1}{c}{\bf \itlang}\\
    \midrule    
    
    \multirow{3}{*}{\bf Train} & Instances
    & 
    \num{5428} &
    \multicolumn{1}{c}{--} &
    \multicolumn{1}{c}{--} &
    \multicolumn{1}{c}{--} &
    \multicolumn{1}{c}{--} &
    \multicolumn{1}{c}{--} &
    %\num{2738} & 
    \multicolumn{1}{c}{--} &
    \multicolumn{1}{c}{--} &
    \multicolumn{1}{c}{--} &
    \multicolumn{1}{c}{--} &
    &
    \num{48042} & 
    \num{39428} &
    \num{1144}\\
    
    & 
    Unique Words 
    & 
    \num{1265} &
    \multicolumn{1}{c}{--} &
    \multicolumn{1}{c}{--} &
    \multicolumn{1}{c}{--} &
    \multicolumn{1}{c}{--} & 
    \multicolumn{1}{c}{--} &
    %\num{871} & 
    \multicolumn{1}{c}{--} &
    \multicolumn{1}{c}{--} &
    \multicolumn{1}{c}{--} &
    \multicolumn{1}{c}{--} &
    & 
    \num{23213} &
    \num{20221} & 
    \num{721}\\
    
    & Avg. Context Len   & 16.8 &
    \multicolumn{1}{c}{--} &
    \multicolumn{1}{c}{--} &
    \multicolumn{1}{c}{--} &
    \multicolumn{1}{c}{--} &
    \multicolumn{1}{c}{--} &
    \multicolumn{1}{c}{--} &
    \multicolumn{1}{c}{--} &
    \multicolumn{1}{c}{--} &
    \multicolumn{1}{c}{--} &
    & 32.7 & 32.3 & 23.2\\
    \midrule
    \multirow{3}{*}{\bf Dev} & 
    Instances &
    \num{638} &
    \num{998} & 
    \num{852} & 
    \num{98} & 
    \num{200} & 
    \num{104} &
    %\num{66} & 
    \num{208} & 
    \num{404} & 
    \num{250} & 
    \num{3046} &
    & 
    \num{8870} &
    \num{8588} &
    \num{198}\\
    
    & 
    Unique Words &
    \num{599} & 
    \num{354} & 
    \num{542} & 
    \num{63} & 
    \num{174} & 
    \num{82} &
    %\num{20} & 
    \num{137} & 
    \num{183} & 
    \num{150} & 
    \num{867} &
    & 
    \num{4383} & 
    \num{3517} & 
    \num{136}\\
    & 
    Avg. Context Len & 17.1& 8.4 & 32.6 & 19.8&24.8 &17.9 &20.8 &5.7 &20.1 &45.7& &32.5 & 34.3 &23.2\\
     \midrule
    \multirow{3}{*}{\bf Test} & 
    Instances 
    & 
    \num{1400} &
    \num{1220} & 
    \num{3406} & 
    \num{390} &
    \num{800} &
    \num{408} & 
    %\num{260} & 
    \num{824} & 
    \num{1014} & 
    \num{1004} & 
    \num{5538} &
    &
    \num{24268} & 
    \num{22232} & 
    \num{592}\\
    & 
    Unique Words 
    & 
    \num{1184} & 
    \num{567} &
    \num{2088} & 
    \num{276} &
    \num{533} &
    \num{305} &
    %\num{105} & 
    \num{476} & 
    \num{475} & 
    \num{600} & 
    \num{1888} &
    & 
    \num{11734} & 
    \num{3517} & 
    \num{394}\\
    & Avg. Context Len  & 17.2 & 8.5& 32.6 &19.4 & 23.5 & 18.1 & 20.6& 6.0&19.9 &46.0 & & 32.9& 36.4& 23.4\\
    \bottomrule
    \end{tabular}
    }
    \caption{Statistics for WordNet and Wiktionary datasets for different languages.} %As for WordNet training data, we report the statistics of our automatically built datasets. {\color{red}Isn't it the case for Wiktionary as well?} \color{red}True i meant the one we created with automatic translation. But we said to remove them anyway from here.}
    \label{tab:stats}
\end{table*}

\subsubsection{Multilingual WordNet}
\label{multilingual-wordnet}

WordNet \cite{miller1995wordnet} is the \textit{de facto} sense inventory for English WSD. 
The resource was originally built as an English lexical database in 1995, but since then there have been many efforts to extend it to other languages \cite{bond2012survey}. 
We took advantage of these extensions to construct XL-WiC. In particular, we processed the WordNet versions of 
Bulgarian \cite{simovandpetya-2010-wordnet-bg}, 
Chinese \cite{huangetal-2014-wordnet-zh}, 
Croatian \cite{raffaelli-etal-2008-wordnet-hr},
Danish \cite{pedersen-et-al-2009-wordnet-da}, 
Dutch \cite{postma-et-al-2016-wordnet-nl}, 
Estonian \cite{vider-and-orav-2002-wordnet-et}, 
Japanese \cite{isahara-et-al-2008-wordnet-ja}, 
Korean \cite{yoon-et-al-2009-wordnet-ko} and
Farsi \cite{Shamsfard2009SemiAD}.\footnote{We tried other WordNet versions such as Albanian \cite{ruci:2008-albanet}, 
Basque \cite{pociello-etal-2008-wnterm}, Catalan \cite{benitez1998methods}, Galician \cite{guinovart2011galnet}, Hungarian \cite{mihaltz-et-al-2008-wordnet-hu}, Italian \cite{pianta-etal-2002-wordnet-it}, Slovenian \cite{fiser-etal-2012-wordnet-sl} and Spanish \cite{atserias-et-al-2004-wordnet-es}; but, they did not contain enough examples.}

%\red{complete XXX - missing Croatian+Italian}
%\red{Enumerate and cite all language-specific wordnets that we use - Alessandro?}

\paragraph{Farsi: Semi-automatic extraction. }
\label{sec:farsi-dataset}
%We used FarsNet \cite{Shamsfard2009SemiAD}, the largest available WordNet for the Farsi language, for the construction of the corresponding dataset.
FarsNet v3.0 \cite{Shamsfard2009SemiAD} comprises 30K synsets with over 100K word entries.
Many of these synsets are mapped to the English database; however, each synset provides just one example usage for a target word.
This prevents us from applying the automatic extraction of positive examples.
Therefore, we utilized a semi-automatic procedure for the construction of the Farsi set. To this end, for each word, we extracted all example usages from FarsNet, and asked an annotator to group them into positive and negative pairs.
The emphasis was to make a challenging dataset with sense distinctions that are easily interpretable by humans.
This can also be viewed as a case study to understand the real gap between human and machine performance in settings where manual curation of instances is feasible. %For this specific dataset, three checker annotators (all Farsi native speakers) were asked to independently label a subset of the dataset (see Section \ref{sec:validation}).
%\footnote{The checker annotators were given a very brief introduction of the task, with 10 separate labeled instances.}
%The average agreement of each with the original labels was 97\% (in terms of accuracy).\footnote{
%An adjudication of the annotations would have further raised the bar given that some of the disagreements were due to careless mistakes by the checker annotators.}
%
\paragraph{Filtering.} WordNet is often considered to be a fine-grained resource, especially for verbs \cite{duffield2007criteria}. In some cases, the exact meaning of a word can be hard to assess, even for humans. For example, WordNet lists 29 distinct meanings for the noun \textit{line}, two of which correspond to the horizontally and the vertically organized line formations.
Therefore, to cope with this issue, we followed \citet{pilehvar2019-wic} and filtered out all pairs whose target senses were connected by an edge (including sister-sense relations) in WordNet's semantic network or if they belonged to the same supersense, i.e., one of the 44 lexicographer files\footnote{\url{wordnet.princeton.edu/documentation/lexnames5wn}} in WordNet which cluster concepts into semantic categories, e.g., \textit{Animal}, \textit{Cognition}, \textit{Food}, etc.
\begin{CJK}{UTF8}{}
    \CJKfamily{mj}
For example, the Japanese instance ``成長中の企業は大胆な指導者いなければならない'' (``Growing \textit{companies} must have bold leaders''), ``彼は安定した大きな企業に投資するだけだ'' (``He just invested in big stable \textit{companies}'')
for the target word ``企業'' (``company'') is discarded as its corresponding synsets, i.e., ``An organization created for business ventures'' and ``An institution created to conduct business'', are grouped under the same supersense in WordNet, i.e., \textit{Group}.
\end{CJK}

Finally, all datasets are split into development\footnote{{%maybe move this to experimental setup? 
Development sets are intended to be used for different purposes, such as training or validation, as we will show in our experiments in Section \ref{results:wordnet}.}} and test. 
At the end of this step, we ensure that both test and development sets have the same number of positive and negative instances. An excerpt of examples included in some of our datasets are shown in Table \ref{tab:xlwic_examples}.%\footnote{{Please see supplementary material for more examples in other languages.\red{NO EXAMPLES IN THE SUP MATERIAL YET. SHALL WE REMOVE THIS FOOTNOTE? XXX}}}

\subsubsection{Wiktionary}
\label{wiktionary}

Wiktionary is one of the richest free collaborative lexical databases, available for dozens of languages.
%Another source to compile high-quality examples is Wiktionary.
In this online resource, each word is provided with definitions for its various potential meanings, some of which are paired with example usages. 
However, each language has a specific format, and therefore the compilation of these examples requires a careful language-specific parsing. 
We extracted examples for three European languages for which we did not have WordNet-based data%\footnote{While the goal of this Wiktionary extraction was to get as many language as possible, we also performed a small comparison with the Italian WordNet dataset. %While the size of the WordNet dataset is not large enough to draw reliable conclusions, 
%{\color{red}The supplementary material includes a table comparing the nature of these two datasets (including zero-shot cross-lingual transfer results).}}
, namely French, German, and Italian.\footnote{French WordNet, WoNeF \cite{pradet2014wonef}, is built automatically; German WordNet, GermaNet \cite{hamp1997germanet}, has a very restrictive license; and the Italian WordNet \cite{pianta-etal-2002-wordnet-it} provides too few examples.} Once these examples were compiled, the process to build the final dataset was analogous to that for the WordNet-based datasets (see Section \ref{multilingual-wordnet}), except for the filtering step, which was not feasible as Wiktionary entries are not connected through paradigmatic relations as in WordNet.

For the case of Wiktionary, the number of examples was considerably higher; therefore, we also compiled language-specific training sets, which enabled a comparison between cross-lingual and monolingual models (see Section \ref{results:wiktionary}).
All Wiktionary datasets are split into balanced training, development and test splits, in each of which there are equal number of positive and negative instances.

\subsection{Statistics}

Table \ref{tab:stats} shows the statistics of all datasets, including the total number of instances, unique words, and the context length average.\footnote{We used the multilingual Stanford NLP toolkit, Stanza \cite{qi-et-al-2020-stanza}, with the available pre-trained neural models.} 
Wiktionary-based datasets are substantially larger than the WordNet-based ones, and also provide training sets.
The Chinese datasets feature longer contexts on average and contain the largest number of development and testing instances among WordNet-based datasets. 
Korean, on the other hand, is the one with the shortest contexts, which is expected given its agglutinative nature. %; however, this does not hamper performance as shown in Section \ref{results:wordnet}.
As for the training corpora, German and French datasets contain almost ten times the number of instances in the English training set. 
This allows us to perform a large-scale comparison between cross-lingual and monolingual settings (see Section \ref{results:wiktionary}) as well as a few-shot analysis (Section \ref{sec:fewshot}).

\subsection{Validation and human performance}
\label{sec:validation}

To verify the reliability of the datasets, we carried out manual evaluation for those languages for which we had access to annotators. 
To this end, we presented a set of 100 randomly sampled instances from each dataset to the corresponding annotator in the target language.\footnote{For Farsi, three checker annotators were involved in the validation, each annotating the 100-instance subset. In this case, we report the average accuracy.}
%CR \footnote{For Farsi, we had three checker annotators.}
Annotators were all native speakers of the target language with high-level education. 
They were provided with a minimal guideline: a brief explanation of their task and a few tagged examples. 
We did not provide any lexical resource (or any other detailed instructions) to the annotators with the emphasis to make a challenging dataset with sense distinctions that are easily interpretable to the layman.
Given an instance, i.e., a pair of sentences containing the same target word, their task consisted of tagging it with a \textit{True} or \textit{False} label, depending on the the intended meanings of the word in the two contexts. 

Table \ref{tab:validation} reports human performance for eight datasets in XL-WiC.
All accuracy figures are around 80\%, i.e., in the same ballpark as the original WiC English dataset, which attests the reliability of underlying resources and the construction procedure.
The only exception is for Farsi, for which the checker annotators agree with the gold labels in 97\% of the instances (by average).
This corroborates our emphasis on the annotation procedure for this manually-created dataset to have sense distinctions that are easily interpretable by humans. 
As for Wiktionary, the human agreements are lower than those for the WordNet counterparts.\footnote{Even if not part of XL-WiC, we also compiled and validated a small Italian WordNet dataset to compare it with its Wiktionary counterpart. Table \ref{tab:italian} in Appendix includes an additional table comparing the nature of these two datasets (including zero-shot cross-lingual transfer results).} 
This was partly expected given that the semantic network-based filtering step (see Section \ref{multilingual-wordnet}) was not feasible for the case of Wiktionary datasets due to the nature of the underlying resource.

\begin{table}[t]
    \centering
    \setlength{\tabcolsep}{4.6pt}
    \resizebox{\linewidth}{!}{
    \begin{tabular}{ll lllllll ll}
    \toprule
   % &
   \multicolumn{1}{c}{\textbf{WiC}} & &
      \multicolumn{6}{c}{\textbf{WordNet}} & &
      \multicolumn{2}{c}{\textbf{Wiktionary}}  \\
      \cmidrule(lr){1-1} \cmidrule(lr){3-8} \cmidrule(lr){10-11}
      
      \multicolumn{1}{c}{\bf \en} & &
      \multicolumn{1}{c}{\bf \da} & 
      \multicolumn{1}{c}{\bf \fa} & 
      \multicolumn{1}{c}{\bf \itlang} & 
      %\multicolumn{1}{c}{\bf \hr} & 
      \multicolumn{1}{c}{\bf \ja} & 
      \multicolumn{1}{c}{\bf \ko} &
      \multicolumn{1}{c}{\bf \zh} & &
      \multicolumn{1}{c}{\bf \de} & 
      \multicolumn{1}{c}{\bf \itlang} \\
    \midrule
    %\bf Acc. &
    80.0$^*$ & &
    87.0 &
    97.0 &
    82.0 &
    %85.0 &
    75.0 &
    76.0 &
    85.0 & &
     74.0 &
    78.0 \\
    \bottomrule
    \end{tabular}
    }
    \caption{Human performance (in terms of accuracy) for different languages in XL-WiC. $^*$From the original English WiC dataset.
    }
    \label{tab:validation}
\end{table}
%\newcite{pilehvar2019-wic}

\section{Experimental Setup}
\label{experimentalsetup}
%Zero-shot cross-lingual transfer, few-shot monolingual, etc.

%In this section we describe the common experimental setting for all our experiments.

%\subsection{\red{Comparison systems}}
For our experiments, we implemented a simple, yet effective, baseline based on a Transformer-based text encoder \cite{vaswanietal:17} and a logistic regression classifier, following \citet{wang2019superglue}.
The model takes as input the two contexts and first tokenizes them, splitting the input words into sub-tokens. The encoded representations of the target words are concatenated and fed to the logistic classifier. 
%This latter outputs $0$ or $1$ depending on whether the target word is used in two different meanings or with the same meaning, respectively. 
For those cases where the target word was split by the tokenizer into multiple sub-tokens, we followed %the standard procedure 
\newcite{BERT} and considered the representation of its first sub-token.

%\subsection{\red{Transformer-based language models}}
As regards the text encoder, we carried out the experiments with three different multilingual models, i.e., the multilingual version of BERT \cite{BERT} ({\bf \mbert}) and the base and large versions of XLM-RoBERTa \cite{XLMR} ({\bf \xlmrbase} and {\bf \xlmrlarge}, respectively).
In the monolingual setting, we used the following language-specific models: \bertde\footnote{ \url{huggingface.co/dbmdz/bert-base-german-cased}}, \bertfr  \cite{camembert}\footnote{\url{huggingface.co/camembert/camembert-large}}, \bertit \footnote{\url{huggingface.co/dbmdz/bert-base-italian-xxl-cased}}, and ParsBERT\footnote{\url{github.com/hooshvare/parsbert}} \cite{farahani2020parsbert}, respectively, for German, French, Italian, and Farsi. As for all the other languages covered by the WordNet datasets, i.e., Bulgarian, Chinese, Croatian, Danish, Dutch, Estonian, Japanese and Korean, we used the pre-trained models made available by TurkuNLP.\footnote{\url{github.com/TurkuNLP/wikibert}} We refer to each language-specific model as L-BERT.

%\subsection{Training setup}
In all experiments we trained the baselines to minimize the binary cross-entropy loss between their prediction and the gold label with the Adam \cite{kingma-ba-2015-adam} optimizer. Training is carried out for 10 epochs with the learning rate  fixed to $1e^{-5}$ and weight decay set to $0$. 
%We tuned each model on the development set by selecting the best of the ten training checkpoints. 
As for tuning, results are reported for the best training checkpoint (among the 10 epochs) according to the performance on the development set.%\footnote{We will provide code for easier replication of the results.}

\begin{table*}[!ht]
\centering
\setlength{\tabcolsep}{11.0pt}
\resizebox{\linewidth}{!}{
\begin{tabular}{l ccccccccc}
\toprule
\bf Model  &\bf \bg  &\bf \da &\bf  \et &\bf \fa &\bf \hr&\bf \ja &\bf \ko &\bf \nl & \bf \zh \\
\midrule
\multicolumn{10}{r}{\textit{Zero-shot cross-lingual setting} ~~ \textit{\textbf{Train}: EN ~~--~~ \textbf{Dev}: EN}}\\
\midrule
\mbert &   58.28 & 64.86 & 62.56 & 71.50 &  63.97 & 62.26 & 59.76 & 63.84 & 69.36 \\
\xlmrbase &  60.73 &  64.79 & 62.82 & 69.88 & 62.01 & 60.44 & 66.96 & 65.73 & 65.78 \\
\xlmrlarge &  \textbf{66.48} &  \textbf{71.11} & \textbf{68.71} & \textbf{75.25} & \textbf{72.30} & 63.83 & \textbf{69.63} & \textbf{72.81} & \textbf{73.15} \\
\midrule
\midrule
\multicolumn{10}{r}{
\textit{Test instances translated to English} ~~
\textit{\textbf{Train}: EN ~~--~~ \textbf{Dev}: EN}}\\
\midrule
\mbert &  63.52 & 62.71 & 68.46 & -- & 60.54 & 63.95 & -- & 66.53 & --  \\
\xlmrbase & 60.98 & 60.24 & 62.82 & --& 60.78 & 61.77 & --  & 64.64 & --  \\
\xlmrlarge & 64.43 & 66.64 & 63.84 & --& 69.85 & 64.44 & --  & 72.11 & -- \\
\lbert & 64.02  & 65.38 & 64.62 & -- & 69.61 & \textbf{65.90} & -- & 68.43 & -- \\
\midrule
\multicolumn{10}{r}{
%\textit{Train and Dev innstances translated to Target Language ~~
\textit{\textit{Train and Dev instances translated from English to target language} ~~ \textbf{Train}: T-EN ~~--~~ \textbf{Dev}: T-EN}}\\
\midrule
\mbert      & 56.97 & 60.25 & 59.48  & -- & 66.91 & 58.13 & -- & 60.06 & -- \\
\xlmrbase   & 56.07 & 52.85 & 57.18 & -- & 64.22 & 56.19 & -- & 60.56 & -- \\
\xlmrlarge  & 62.13 & 63.39 & 64.87 & -- & 66.18 & 59.47 & -- & 66.73 & -- \\
\lbert      & 54.26 & 60.57 & 59.49 & -- & 61.52 & 58.98 & -- & 60.46 & -- \\

\bottomrule

\end{tabular}
}
\caption{Results on the WordNet test sets when using only English training data in WiC, either in zero-shot cross-lingual setting (top block) or translation-based settings (the lower two blocks). T-EN is a target language dataset, automatically constructed by translating English instances in WiC. %\red{WILL WE HAVE THE LANGUAGE-SPECIFIC MODELS RESULTS IN THE TRANSLATION SETTINGS AT THE END? I UNDERSTOOD THEY WERE READY? XXXX} %{\color{red}The NMT model used in our experiments did not support Farsi, Korean, and Chinese languages.}
}
\label{tab:wordnet_results_en_train}
\end{table*}

\subsection{Evaluation settings}

We evaluated the baselines with different configuration setups, depending on the data used for training and tuning.
%For our experiments we leveraged the data provided within the \xlwic framework to compare and analyze the results attained by several baseline models in four different settings: 
%\begin{enumerate}
%    \item \textbf{Cross-Lingual Zero-shot Setting}, where training is performed on the English set, fine-tuning is done on the corresponding development sets of English or target languages, and testing is carried out on the target language;
%    \item \textbf{Multilingual Fine-Tuning Setting}, where models are both trained and fine-tuned on the English set and tested in other languages;
 %   \item \textbf{Few-Shot Setting} where models are trained on a few data point in the target language and tested on the same language;
 %   \item \textbf{Monolingual Setting}, where models are trained, tuned and tested on data in the same language;
 %   \item \red{\textbf{Translate... (to confirm)}}

%\end{enumerate}

\paragraph{Cross-Lingual Zero-shot.}
%{\color{red}This better go to experimental setup:}
%In our experiments, we use this English dataset for training purposes (see Section \ref{experimentalsetup}), although XL-WiC provides training sets for few other languages.
This setting aims at assessing the capabilities of multilingual models in transferring knowledge captured in the English language to other languages.
As training set, we used the English training set of WiC.
As for tuning, depending on the setting, we either used the English development set of WiC or language-specific development sets of XL-WiC (Section \ref{multilingual-wordnet}).
%As for testing, 
We report results on all WordNet and Wiktionary datasets of \xlwic, i.e., \alllangs.

\paragraph{Multilingual Fine-Tuning.}
In this setting, models are first trained on the WiC's English training set, and then further fine-tuned on the development sets of the target languages in XL-WiC.
%\footnote{Fine-tuning is done for one epoch only.}
%After training a model on the English dataset, we fine-tune it for \num{1} epoch on the small language-specific dataset.
Depending on the training set used, we report results for two configurations: (i) \textit{EN+Target Language}, combining with WiC's training data and the language-specific WordNet development sets for each language, and (ii) \textit{EN+All Languages}, combining that with all WordNet development sets for all languages in XL-WiC, merged as one dataset.

%\paragraph{Few-Shot Setting.}
% moved to analysis section
%We leverage the datasets created from Wiktionary in German, French and Italian, which allows us to create training data of different sizes. For each language we create $8$ training sets with 10, 25, 50 100, 250, 500, 1000 and all the available instances\footnote{Each increasingly bigger dataset contain the examples of the smaller datasets.}. As development set we used the Wiktionary development set of the target language that is included in \xlwic.

\paragraph{Monolingual.}
In this setting, we trained each model on the corresponding training set of the target language only. 
For the case of WordNet datasets (where no training sets are available), we used the development sets for training. 
In this case we split each development set into two subsets with \num{9}:\num{1} ratio (for training and development).
%, and followed the same training procedure outlined earlier in this Section. 
As for the Wiktionary datasets, we used the corresponding training and development sets for each language (Section \ref{wiktionary}).

\paragraph{Translation.}
In this last setting we make use of existing neural machine translation (NMT) models to translate either the training or the test set, essentially reducing the cross-lingual problem to a monolingual one.
In particular, we used the general-domain translation models from the Opus-MT project\footnote{\url{github.com/Helsinki-NLP/Opus-MT}} \cite{TiedemannThottingal:EAMT2020} available for the following language pairs: English--Bulgarian, English--Croatian, English--Danish, English--Dutch, English--Estonian, and  English--Japanese. The models are trained on all OPUS parallel corpora collection \cite{tiedemann-lrec12}, using 
the state-of-the-art 6-layer Transformer-based architecture \cite{vaswanietal:17}.\footnote{More details about the NMT models and their translation quality are given in the Appendix (Table \ref{tab:BLEU}).}
In this configuration, as the original target word may be lost during automatic translation, we view the task as context (sentence) similarity as proposed by \citet{pilehvar2019-wic}.\footnote{The Appendix (Table \ref{tab:translated_alignements}) includes another translation baseline that uses dictionary alignments to identify the target word as comparison, but performed worse overall.} Therefore, for each model, the context vector is given by the start sentence symbol. We note that while training custom optimized NMT models for each target language may result in better overall performance, this is beyond the scope of this work.

%https://
%In particular, we use \red{(to COMPLETE WITH NMT MODELS USED XXX- ALESSANDRO)}. Given that these models do not provide alignments, we rely on the representation of the sentence (extracted from the given language model) and not that of the target word. Moreover, we include a setting where the target word was aligned across languages. This alignment was performed with a multilingual dictionary \cite[BabelNet]{NavigliPonzetto:12aij} for those training sentences for where an alignment was found.

\paragraph{Evaluation Metrics.}
Since all datasets are balanced, we only report accuracy, i.e., the ratio of correctly predicted instances (true positives or true negatives) to the total number of instances.
\begin{table*}[!ht]
\centering
\setlength{\tabcolsep}{8.0pt}
\resizebox{\linewidth}{!}{
\begin{tabular}{l cccccccccc}
\toprule
\bf Model &\bf \en &\bf \bg  &\bf \da &\bf  \et &\bf \fa &\bf \hr&\bf \ja &\bf \ko &\bf \nl & \bf \zh \\
\midrule

%\multicolumn{11}{r}{\color{magenta}\textit{All instances translated to Target Language + Word Alignment (\textbf{Train}: T-EN ~~-~~ \textbf{Dev}: T-EN)}}\\
%\midrule
%\color{magenta} \mbert & - &60.66 & 60.16& \textbf{61.79}& - & \textbf{68.87} &  52.79 & - & 57.57 & \\
%\color{magenta} \xlmrbase & - & 57.30 & 57.34 & 51.79 & -& 59.80 & 51.70 & - & 60.26 & \\
%\color{magenta} \xlmrlarge & - & \textbf{63.36}  & \textbf{66.27} & 61.54 & - & 66.42& \textbf{53.88} & - &\textbf{69.42} & - \\
%\color{magenta} \lbert & - & 56.31  & 58.07 & 56.67 & - & 59.31& 53.40 & - & 58.47 & - \\
%\midrule

%\midrule
\multicolumn{11}{r}{\textit{\textbf{Train}: EN ~~--~~ \textbf{Dev}: Target Language}}\\
\midrule
\mbert & -- & 59.18 &  64.59 &  63.08 & 70.38 &64.95 & 59.95 & 63.31 & 64.04 & 70.48 \\
\xlmrbase & -- &60.74  & 64.80 &  60.77 & 67.75& 62.50 & 57.65 & 66.96 & 61.85 & 65.78 \\
\xlmrlarge & -- &\textbf{66.48} & \textbf{71.10}  &  \textbf{68.72} & \textbf{73.63} & \textbf{72.30} & \textbf{60.92} & \textbf{69.63} & \textbf{69.62} & \textbf{73.15} \\
\midrule
\multicolumn{11}{r}{\textit{\textbf{Train}: EN+Target Language ~~--~~ \textbf{Dev}: EN}}\\
\midrule
\mbert & -- & 71.72  & 62.62 & 63.08 & 69.38 & \textbf{72.30} &  60.92 & 70.91 & 62.95 & 76.72 \\
\xlmrbase & -- & 64.51  & 64.45 &  60.00 & 65.38& 71.57 & 58.37 & 65.68 & 64.54 & 73.46 \\
\xlmrlarge & -- & \textbf{75.41}  & \textbf{70.52} & \textbf{68.97} & \textbf{73.75}& 69.61 & \textbf{63.11} & \textbf{73.47} & \textbf{74.50} & \textbf{77.52} \\
\midrule
\multicolumn{11}{r}{\textit{\textbf{Train}: EN+All Languages ~~--~~ \textbf{Dev}: EN}}\\
\midrule
\mbert & -- &73.03  & 65.09 & 62.31 &73.63& 72.30 &  65.53 & 71.01& 67.73 &76.53 \\
\xlmrbase & -- &67.30  &67.62 & 59.49 &64.50&  66.18 & 57.77 &67.06& 66.33 &71.02 \\
\xlmrlarge & -- &\textbf{78.44}  & \textbf{71.49} &  \textbf{72.05} & \textbf{78.25} & \textbf{76.96}& \textbf{66.38} &\textbf{76.53}& \textbf{77.49}&\textbf{78.95} \\
\midrule
\multicolumn{11}{r}{\textit{\textbf{Train}: Target Language ~~--~~ \textbf{Dev}: Target Language}}\\
\midrule
\mbert & 66.71  & \textbf{82.30} & 62.13 & 58.21 & 63.75 & 77.45 & \textbf{61.04} &70.71 & 64.84 & 76.09\\
\xlmrbase & 64.36 & 79.75  & 64.00 & \textbf{64.36} & 66.25& \textbf{79.17} & 58.86 &70.61 &66.33 &78.11 \\
\xlmrlarge & 70.14  & 82.05  & \textbf{66.53} & 59.23 & 68.00&76.72 &  55.22 & \textbf{73.08} & \textbf{69.42} & \textbf{81.83}\\
\lbert & 69.60 & 81.23  &62.60 & 58.46 & \textbf{76.63} &76.47 & 56.07 & 58.68 & 68.73 & 77.36 \\
%$\,$69.90$^\dagger$^{\dagger\mathbin{\Diamond}}
\bottomrule

\end{tabular}
}
\caption{Results on the WordNet test sets when using language-specific data, either for training or for tuning.}
\label{tab:wordnet_results_language_spec}
\end{table*}

\section{Results}
In this section, we report the results for the configurations discussed in the previous section on the \xlwic benchmark.
We organize the experiments into two parts, based on the test dataset: WordNet (Section \ref{results:wordnet}) and Wiktionary (Section \ref{results:wiktionary}).

%The results of our XL-WiC experiments are presented in Section \ref{results:wordnet} (WordNet) and \ref{results:wiktionary} (Wiktionary). 

\subsection{WordNet datasets}
\label{results:wordnet}

\paragraph{Using English data only.}
Table \ref{tab:wordnet_results_en_train} shows results on the XL-WiC WordNet test sets, when only WiC's English data was used for training and tuning purposes. 
Across the board, \xlmrlarge consistently achieves the best results, while \mbert and \xlmrbase attain scores in the same ballpark. 
Indeed, the massive pretraining and the number of parameters of \xlmrlarge play key roles behind this lead in performance.
As regards the translation-based settings (lower two blocks), the performance generally falls slightly behind the zero-shot cross-lingual counterpart. 
%The translation quality may impact the results, 
This shows that the usage of good quality English data and multilingual models provide a stronger training signal than noisy automatically-translated data. This somehow contrasts with the observations made on other cross-lingual tasks in XTREME \cite{hu2020xtreme}, especially in question answering datasets \cite{artetxe2019cross,lewis2020mlqa,clark2020tydi}, where translating data was generally better.
This difference could perhaps be reduced with larger monolingual models or accurate alignment, but this would further increase the complexity, and extracting these alignments from NMT models is not trivial \cite{koehn-knowles-2017-six,ghader-monz-2017-attention,li-etal-2019-word}.

%In Table \ref{tab:wordnet_results_en_train} and \ref{tab:wordnet_results_language_spec} we report the results on the WordNet test sets for several settings.\footnote{In this table we do not include the results on the WordNet Italian corpora as its development and test sets were too small to draw reliable conclusions. Nonetheless, zero-shot cross-lingual transfer results for Italian can be found in the supplementary material.} 
%In the zero-shot setting (top block in Table \ref{tab:wordnet_results_en_train}), results are consistent across languages with \xlmrlarge model attaining the best results that range between \num{63.84} accuracy in \jalong to {\color{red}\num{81.54} in \itlanglong}.

\paragraph{Utilizing language-specific data.}
Table \ref{tab:wordnet_results_language_spec} shows results for settings where target language-specific data was used for training or tuning.
Comparing the results in the top block (where target language data was used for tuning) with the middle two blocks (where target language data was instead used for training) reveals that it is more effective to leverage the target language data for training, rather than using it for tuning only.
Overall, it is clear that adding multilingual data during training drastically improves the results in all languages.
In this case, training (fine-tuning) is performed on a larger dataset which, despite having examples from different languages, provides a stronger signal to the models, enabling them to better generalize across languages. 
On the contrary, when only using target language data for training and tuning (last block in the table), results drop for most languages.
This highlights the fact that having additional training data is beneficial, reinforcing the utility of multilingual models and cross-lingual transfer.

%{\color{red}Interestingly, when target language development sets are used fine-tuning (top block in {\bf }Table \ref{tab:wordnet_results_language_spec}), models' results improve on 6 languages out of 10 (not considering English).} Even better results are achieved by fine-tuning the models on the development sets of all languages together (Train: EN + All Languages). This, indeed, dramatically increases the baselines' performance on all languages. 

\begin{table}
\setlength{\tabcolsep}{9.0pt}
\centering
\resizebox{\columnwidth}{!}{
\begin{tabular}{cl ccc}
\toprule
%\bf Setting 
&\bf  Model &\bf  \de &\bf  \fr &\bf  \itlang   \\
\midrule

\multirow{3}{*}{\rotatebox[origin=c]{90}{\bf Z-Shot}}

& \mbert & 58.27 &56.00 &58.61  \\
& \xlmrbase & 58.30 &56.13 & 55.91  \\
& \xlmrlarge &\textbf{65.83} & \textbf{62.50} & \textbf{64.86} \\
\midrule

\multirow{4}{*}{\rotatebox[origin=c]{90}{\bf Mono}}

& \mbert & 81.58 & 73.67 & 71.96  \\
& \xlmrbase &80.84  & 73.06 & 68.58  \\
& \xlmrlarge & \textbf{84.03}& 76.16 &72.30   \\
& L-BERT & 82.90 & \textbf{78.14} & \textbf{72.64}  \\
\bottomrule
\end{tabular}
}
\caption{Results on the Wiktionary test sets in different training settings: zero-shot (Z-Shot) and monolingual training (Mono). L-BERT stands for language-specific models, i.e.,  \bertde, \bertfr and \bertit for German, French and Italian, respectively.  
}
\label{tab:wiktionary_results}
\end{table}

\subsection{Wiktionary Datasets}
\label{results:wiktionary}

In Table \ref{tab:wiktionary_results} we show results for the Wiktionary datasets. 
Differently from the results reported for the WordNet datasets (Table \ref{tab:wordnet_results_en_train}), models are less effective in the zero-shot setting, performing from \num{10} to almost \num{20} points lower than their counterparts trained on data in the target language.
This can be attributed to the size of the available training data. 
Indeed, while in the WordNet datasets we only have a very small amount of data at our disposal for training (see statistics in Table \ref{tab:stats}), Wiktionary training sets are much larger, hence providing enough data to the models to better generalize. 
Once again, \xlmrlarge proves to be the best model in the zero-shot setting and a competitive alternative to the language-specific models (\lbert  row) in the monolingual setting, performing \num{1.1} points higher in German and \num{2} and \num{0.3} lower in French and Italian, respectively.

\section{Analysis}

In this section, we delve into the performance of the models on \xlwic and analyze relevant aspects about their behaviour.

\subsection{Seen and Unseen Words}
For this analysis, we aim at measuring the difference in performance when a given target word was seen (as a target word) at training time or not. 
To this end, we evaluate our baselines when trained on the German, French, and Italian Wiktionary training sets and tested on two different subsets of the larger language-specific Wiktionary test sets: In-Vocabulary (\textbf{IV}), containing only the examples whose target word was seen at training time; and Out-of-Vocabulary (\textbf{OOV}), containing only the examples whose target word was not seen during training. We report the results in Table \ref{tab:oov}.

\begin{table}[t]
\centering
\setlength{\tabcolsep}{9.0pt}
\resizebox{\columnwidth}{!}{
\begin{tabular}{cl ccc}
\toprule
\bf % Setting 
&\bf  Model &\bf  \de & \bf \fr &\bf  \itlang \\
\midrule

\multirow{4}{*}{\rotatebox[origin=c]{90}{\bf IV}}

& \mbert & 81.86 & 72.92 & 73.15  \\
& \xlmrbase & 81.17 & 71.92 & 70.69 \\
& \xlmrlarge & \textbf{84.24} & 75.61 & \textbf{75.12} \\
& L-BERT & 83.23 & \textbf{77.62} & 73.89\\

\midrule

\multirow{4}{*}{\rotatebox[origin=c]{90}{\bf OOV}}
& \mbert & 70.08 & 71.24 & 68.54  \\
& \xlmrbase & 71.31 & 71.14 & 62.36 \\
& \xlmrlarge & 72.54 & 73.93 & 65.17 \\
& L-BERT & \textbf{76.64} & \textbf{78.00} & \textbf{69.10}\\

\bottomrule
\end{tabular}
}
\caption{Results on the in-vocabulary ({\bf IV}) and out-of-vocabulary ({\bf OOV}) Wiktionary test sets. L-BERT stands for each language-specific model, i.e.,  BERT-base-de, camemBERT and BERT-base-xxl-it for German, French and Italian, respectively.  
}
\label{tab:oov}
\end{table}
\begin{figure*}
        \includegraphics[scale=0.40]{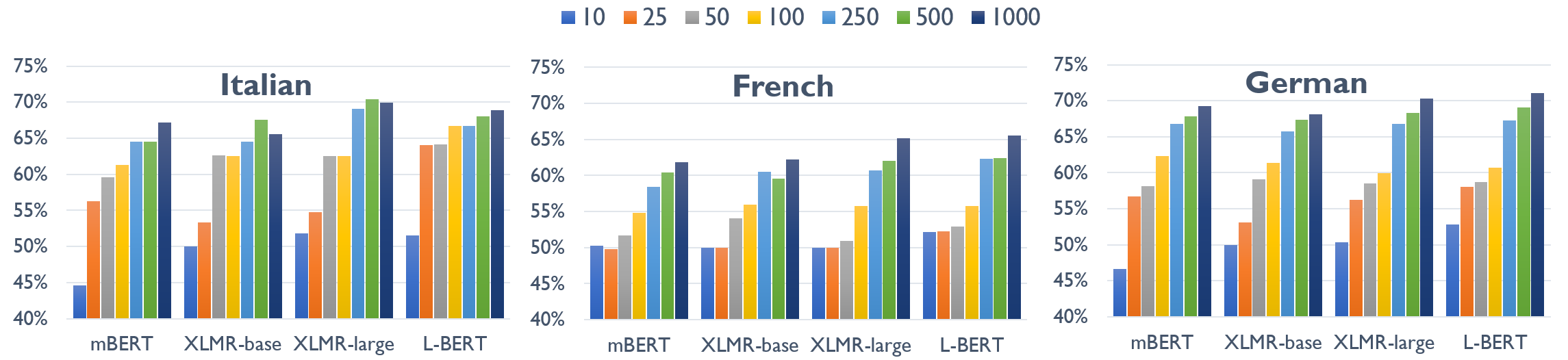}
        \caption{\label{fig:few-shot} The impact of training set size (\# of instances) on performance, for the Wiktionary datasets.}
\end{figure*}

In general, multilingual models are less reliable when classifying unseen instances, lagging between 1 and \num{12} points behind in performance, depending on the language and on the model considered. This can be attributed to the fact that their vocabulary is shared among several languages, and therefore may have less knowledge stored about particular words that do not occur often. The performance drop of language-specific models (\lbert) is less pronounced, with the French architecture (\bertfr) attaining even higher performance (\num{0.4} points more) on the OOV set. 
%We leave careful investigation and analysis of these observations to future work.
%This might be due to the fact that \bertfr is larger than the German and Italian models (3x times their parameters) hence being able to provide more consistent representations.  

%will show it in the supplementary material, few-shot fixed size is the main experiment that will remain on the paper
%\subsection{Training size}
%\red{Add one figure for each language with the \% of training data size}

\subsection{Few-shot Monolingual}
\label{sec:fewshot}
As an additional experiment, we investigate the impact of training size on performance.
To this end, we leveraged the Wiktionary datasets for German, French and Italian, which allow us to use varying-sized training sets, and created $7$ training sets with 10, 25, 50, 100, 250, 500, 1000 instances.\footnote{Bigger datasets are supersets of smaller ones.} %As development set we used the Wiktionary development data in the corresponding target language.

%In Figure \ref{fig:few-shot} we report the performance of the baseline models in this few-shot experiment on the Wiktionary datasets. 
The results of this experiment are displayed in Figure \ref{fig:few-shot}.
When providing only 10 examples, most of the models perform similarly or even worse than random, i.e., $50\%$ accuracy. 
In this setting, language-specific models (\lbert) attain better results than their multilingual counterparts, showing better generalization capabilities when fewer examples are provided. This also goes in line with what we found in the previous experiment on seen and unseen words.
With less than $5\%$ of the training data (1000 instances in French and German and 50 instances in Italian), all models attain roughly $85\%$ of their performance with full training data, comparable to results reported for the \zeroshot setting (Table \ref{tab:wiktionary_results}).

\section{Conclusions}

In this paper we have introduced XL-WiC, a large benchmark for evaluating context-sensitive models. 
XL-WiC comprises datasets for a heterogeneous set of 13 languages, including the original English data in WiC \cite{pilehvar2019-wic}, providing an evaluation framework not only for contextualized models in those languages, but also for experimentation in a cross-lingual transfer setting.  
Our evaluations show that, even though current language models are effective performers in the zero-shot cross-lingual setting (where no instances in the target language are provided), there is still room for improvement, especially for far languages such as Japanese or Korean.

As for future work, we plan to investigate using languages other than English for training (e.g., our larger French and German training sets) in our cross-lingual transfer experiments, since English may not always be the optimal source language \cite{anastasopoulos2020should}.
%We facilitate this research by providing large training sets for three more languages (French, German and Italian) and additional validation tests for the remaining languages, i.e., \wnlangs.
%\red{Manual filtering as Farsi... test in a setting fully interpretable by humans...??}
Finally, while in our comparative analysis we have focused on a quantitative evaluation for all languages, an additional 
%more carefully 
error analysis per language would be beneficial in revealing the weaknesses and limitations of cross-lingual models.
%{\color{orange}We facilitate this further analysis by sharing all the individual outputs and models considered in the evaluation in addition to the entire XL-WiC benchmark.}

%\end{spacing}

\section*{Acknowledgments}

We would like to thank Qincheng Zhang (Chinese), Janine Siewert (Danish), Houman Mehrafarin, Hossein Mohebbi, and Ali Modarresi (Farsi), \'{A}ngela Collados A\'{i}s (German), Asahi Ushio (Japanese), and Yunseo Joung (Korean) for their help with the manual evaluation of the datasets. 
 
\begin{comment}
\vspace{1ex}
\noindent
\begin{minipage}{0.1\linewidth}
    \vspace{-10pt}
    \raisebox{-0.2\height}{\includegraphics[trim =32mm 55mm 30mm 5mm, clip, scale=0.2]{erc.png}} \\
    \raisebox{-0.25\height}{\includegraphics[trim =0mm 5mm 5mm 2mm,clip,scale=0.078]{eu.pdf}}
\end{minipage}
\hspace{0.01\linewidth}
\begin{minipage}{0.85\linewidth}
Alessandro gratefully acknowledges the support of the FoTran project, funded by the European Research Council (ERC) under the European Union's Horizon 2020 research and innovation programme (grant agreement No 771113), and the CSC – IT Center for Science, Finland, for computational resources.

Tommaso gratefully acknowledges the support of the ERC Consolidator Grant MOUSSE No. 726487 under the European Union’s Horizon 2020 research and innovation programme.
  \vspace{1ex}
  \end{minipage}
\end{comment}
  \noindent
\begin{figure}[!ht]
\begin{subfigure}{0.1\columnwidth}
    \raisebox{-0.2\height}{\includegraphics[trim =32mm 55mm 30mm 5mm, clip, scale=0.2]{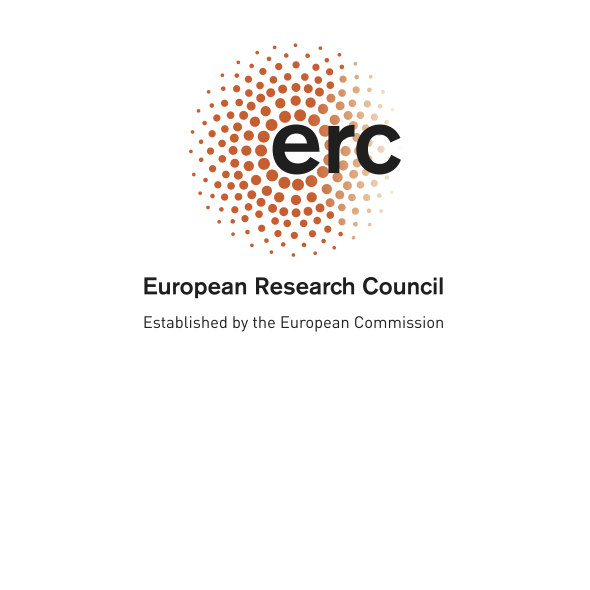}}
\end{subfigure}
\hspace{0.01\linewidth}
\begin{subfigure}{0.72\columnwidth}
  Alessandro gratefully acknowledges the support of the FoTran project, funded by the European Research Council (ERC) under the European Union's Horizon 2020 research and innovation programme (grant agreement No 771113), and the CSC – IT Center for Science, Finland, for computational resources.
    \\
    
    Tommaso gratefully acknowledges the support of the ERC Consolidator Grant MOUSSE No. 726487 under the European Union’s Horizon 2020 research and innovation programme.
\end{subfigure}
\hspace{0.01\linewidth}
\begin{subfigure}{0.05\columnwidth}
\raisebox{-0.25\height}{\includegraphics[trim =0mm 5mm 5mm 2mm,clip,scale=0.078]{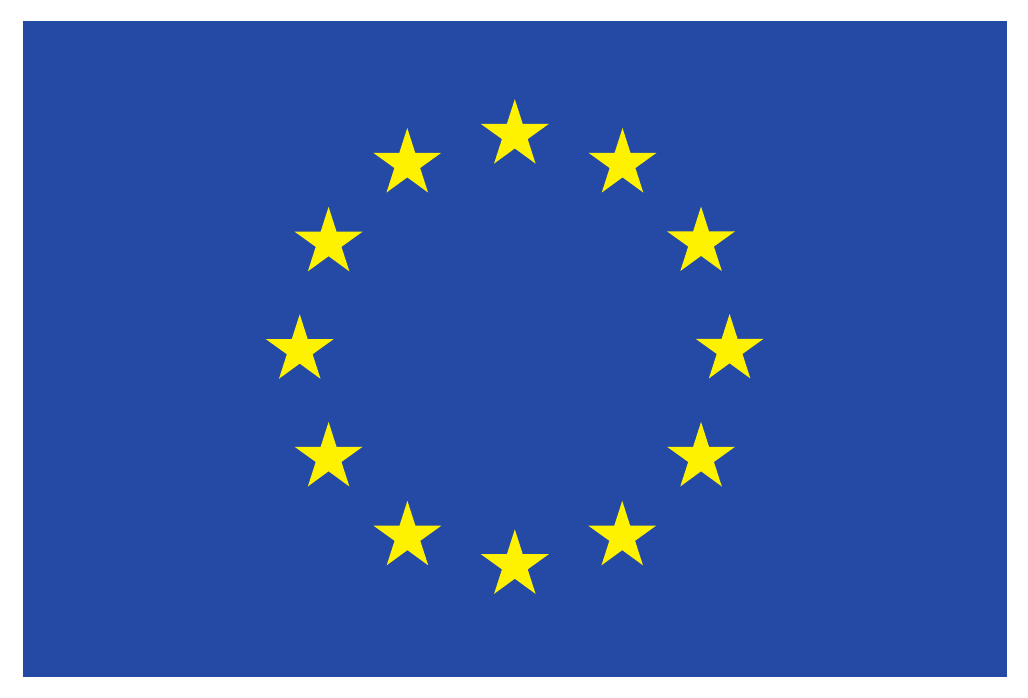}}
\end{subfigure}
\end{figure}

\bibliography{emnlp2020}
\bibliographystyle{acl_natbib}

\appendix

\section*{Appendix}
%\label{sec:appendix}

%Appendices are material that can be read, and include lemmas, formulas, proofs, and tables that are not critical to the reading and understanding of the paper. 
%Appendices should be \textbf{uploaded as supplementary material} when submitting the paper for review.
%Upon acceptance, the appendices come after the references, as shown here.

\section{Comparison between Italian WordNet and Wiktionary datasets}

In Table \ref{tab:italian} we provide a small comparison between the Italian WordNet and Wiktionary datasets, which is the only language that overlaps. While this comparison is quite limited, it provides a few hints on the qualitative differences between Wiktionary and WordNet datasets.

\begin{table}[ht]
    \centering
    \begin{tabular}{llrr}
    \toprule
         && WordNet & Wiktionary\\
        \midrule
 \multirow{3}{*}{\rotatebox[origin=c]{90}{Stats}}
 & Instances & 260 & 592 \\
 & Unique Words & 105 & 394 \\
 & Avg. Ctx length & 14.53 &  23.39 \\
 & Human acc. & 82.0 & 78.0 \\
 \midrule
\multirow{3}{*}{\rotatebox[origin=c]{90}{Z-shot}} & XLM-R base & 66.15 & 55.91 \\
 &XLM-R large & 80.00 & 64.86 \\
 &mBERT & 70.00 & 58.61 \\
 \bottomrule
    \end{tabular}
    \caption{Statistics and comparison between the Italian WordNet and the Italian Wiktionary WiC datasets. Zero-shot results are computed by using the original English WiC \cite{pilehvar2019-wic} for training and development.}
    \label{tab:italian}
\end{table}

\section{Models' Parameters}
In Table \ref{tab:mod_params} we report the parameters of the models used in our experiments.
\begin{table}[ht]
    \centering
    \begin{tabular}{lrr}
    Model & Trainable Parameters \\
    \toprule
         \mbert &  110M \\
         BERT-large & 335M\\
         \xlmrbase & 270M \\
         \xlmrlarge & 550M \\
         \bottomrule
    \end{tabular}
    \caption{Number of parameters for our comparison systems.}
    \label{tab:mod_params}
\end{table}

\section{Additional experimental results}

\paragraph{WordNet datasets.}
Table \ref{tab:zshot-wnet} includes details on the variability of the results, in particular the average results from three runs, including the standard deviation, for the zero-shot cross-lingual setting - this is the setting producing a higher variability in the results.

%\cite{pilehvar2019-wic}

\paragraph{Wiktionary Datasets.}
In Table \ref{tab:zshot-wikt} we show the development and test results in the monolingual settings of the multilingual language models trained, tuned and tested on the \xlwic language-specific datasets from Wiktionary. 

\paragraph{Translation setting + Dictionary alignment.}
%In particular, given that these models do not provide alignments, we rely on the representation of the sentence (extracted from the given language model) and not that of the target word.
We include a setting where, after translating the English training set to each target language, we also retrieve the corresponding translation of the English target word through a multilingual dictionary. We use BabelNet \cite{NavigliPonzetto:12aij} as multilingual dictionary for all languages, discarding the sentences where the translated target word could not be found. Table \ref{tab:translated_alignements} shows the results.

\section{Translation models}
Translation models are trained using the MarianNMT framework \cite{mariannmt} on a filtered version of all OPUS parallel corpora collection using a language identifier (CLD2). As hyper-parameters, each model is based on the \textit{base} version of the Transformer architecture \cite{vaswanietal:17}. All models and training details are available at \url{https://github.com/Helsinki-NLP/Opus-MT}. To give an idea of the translation quality, Table \ref{tab:BLEU} reports the BLEU scores \cite{papineni-etal-2002-bleu} for each model.  We report the performance, as described within the Opus-MT project, on the latest available test sets from the series of WMT news translation shared tasks, or on 5K sentences taken from either the Tatoeba corpus \cite{tiedemann-lrec12}, or the Bible corpus \cite{christodouloupoulos2015massively}: 
\begin{compactitem}
    \item Bulgarian (BG): Tatoeba, model checkpoint EN$\leftrightarrow$BG \textit{opus-2019-12-18}
    \item Danish  (DA): Tatoeba, model checkpoint EN$\leftrightarrow$DA \textit{opus-2019-12-18}
    \item Estonian (ET): newstest2018, model checkpoint EN$\leftrightarrow$ET \textit{opus-2019-12-18}
    \item Croatian (HR): Tatoeba, model checkpoint EN$\to$HR \textit{opus-2019-12-04}, HR$\to$EN \textit{opus-2019-12-05}
    \item Japanese (JA): bible-uedin, model checkpoint EN$\to$JA \textit{opus-2020-01-08}, JA$\to$EN \textit{opus-2019-12-18}
    \item Dutch (NL): Tatoeba, model checkpoint EN$\to$NL \textit{opus-2019-12-04}, NL$\to$EN \textit{opus-2019-12-05}
\end{compactitem}

\begin{table*}[t]
\setlength{\tabcolsep}{10.0pt}
%\scalebox{0.9}{
%\resizebox{\columnwidth}{!}{%
%\small
%\resizebox{\linewidth}{!}{
\centering
\begin{tabular}{l ccc r}
\toprule

Language & \mbert & \xlmrbase & \xlmrlarge & Size\\
\cmidrule(lr){1-1} \cmidrule(lr){2-2} \cmidrule(lr){3-3} \cmidrule(lr){4-4} \cmidrule(lr){5-5}
%\en & $68.70 \pm 0.50$ & $68.39 \pm 0.86$  & $\textbf{73.77} \pm 0.48$ & -\\
%\midrule
\bg & $58.72 \pm 1.60$ & $58.42 \pm 3.74$& $\textbf{64.70} \pm 2.05$ & \num{1220}\\
\da & $63.26 \pm 2.25$ & $62.29 \pm 2.34$ & $\textbf{68.91} \pm 2.45$ & \num{3406}\\
\et & $59.40 \pm 2.74$ & $62.74 \pm 2.43$& $\textbf{68.29} \pm 2.97$ & \num{390}\\
\fa & $67.71 \pm 3.36$ & $64.67 \pm 5.07$& $\textbf{73.58} \pm 1.49$ & 
\num{800}\\
\hr & $65.93 \pm 2.25$ & $63.07 \pm 2.06$& $\textbf{68.63} \pm 3.20$ & \num{408}\\
%\itlang & $70.26 \pm 1.18$ & $68.08\pm 1.92$& $\textbf{79.10} \pm 2.99$ & \num{260}\\
\ja & $62.58 \pm 0.37$ & $59.63 \pm 3.11$& $\textbf{63.23} \pm 1.72$ & \num{824}\\
\ko & $61.21 \pm 2.34$ & $64.04 \pm 3.93$ & $\textbf{69.40} \pm 3.61$ & \num{1014}\\
\nl & $63.55 \pm 1.37$ & $64.41 \pm 1.40$ & $\textbf{72.14}\pm 1.88$ & \num{1004}\\
\zh & $68.85 \pm 0.50$ & $61.69 \pm 3.78$& $\textbf{70.68} \pm 2.94$ & \num{5538}\\

\bottomrule
\end{tabular}
%}
\caption{Zero-shot results on \mbert, \xlmrbase and \xlmrlarge on the WordNet-based datasets when using the English WiC training and development sets. }
\label{tab:zshot-wnet}
\end{table*}

\begin{table*}[t]
\setlength{\tabcolsep}{7.0pt}
\centering
%\scalebox{0.9}{
%\resizebox{\linewidth}{!}{%
%\small
\begin{tabular}{l cccccc rrr}
\toprule
& \multicolumn{2}{c}{\mbert} & \multicolumn{2}{c}{\xlmrbase} & \multicolumn{2}{c}{\xlmrlarge} &  \multicolumn{3}{c}{Size} \\
Language & Dev & Test & Dev & Test & Dev & Test & Train & Dev & Test \\
\cmidrule(lr){1-1} \cmidrule(lr){2-3} \cmidrule(lr){4-5} \cmidrule(lr){6-7} \cmidrule(lr){8-10}

\de  & 79.73 & 81.58 & 78.93 & 80.84 & 83.03 & \textbf{84.03}&\num{48042}& \num{8870}&\num{24268}\\
\fr & 71.62& 73.67 & 71.43& 73.06& 75.00 & \textbf{76.16} &\num{39428} & \num{8588} & \num{22232}\\
\itlang  &73.23 & 71.96 & 75.25 & 68.58 & 74.24 & \textbf{72.30}& \num{1144} & \num{198} & \num{592}\\

\bottomrule
\end{tabular}
%}
\caption{Results on \mbert, \xlmrbase and \xlmrlarge on the Wiktionary-based datasets when using the language-specific training and development data.} %Notation is similar to Table \ref{tab:evCoarseWSD}.} %FT stands for Fine-tune.}
\label{tab:zshot-wikt}
\end{table*}

\begin{table*}[t]
    \centering
\begin{tabular}{l cccccccccc}
\toprule
\bf Model &\bf \en &\bf \bg  &\bf \da &\bf  \et &\bf \fa &\bf \hr&\bf \ja &\bf \ko &\bf \nl & \bf \zh \\
\midrule    
\multicolumn{11}{r}{\textit{All instances translated to Target Language + Dictionary Alignment (\textbf{Train}: T-EN ~~-~~ \textbf{Dev}: T-EN)}}\\
\midrule
\mbert & - &60.66 & 60.16& \textbf{61.79}& - & \textbf{68.87} &  52.79 & - & 57.57 & - \\
\xlmrbase & - & 57.30 & 57.34 & 51.79 & -& 59.80 & 51.70 & - & 60.26 & - \\
\xlmrlarge & - & \textbf{63.36}  & \textbf{66.27} & 61.54 & - & 66.42& \textbf{53.88} & - &\textbf{69.42} & - \\
\lbert & - & 56.31  & 58.07 & 56.67 & - & 59.31& 53.40 & - & 58.47 & - \\
    \bottomrule
    \end{tabular}
    \caption{Results on the WordNet test sets when using automatically-translated data with a multilingual dictionary-alignment technique.}
    \label{tab:translated_alignements}
\end{table*}
% based on the sense inventory

\begin{table*}[t]
\centering
% \resizebox{\linewidth}{!}{
\begin{tabular}{lcccccc}
\toprule
 Opus-MT & \bg & \da  & \et & \hr  & \ja  & \nl  \\ 
\midrule
EN$\to$XX & 50.0 &  60.4  & 23.3  & 48.3  & 42.1  & 57.1  \\ 
XX$\to$EN & 59.4 & 63.6  & 30.3  & 58.7  & 41.7  & 60.9  \\ 
\bottomrule
\end{tabular}
%}
\caption{BLEU score of the translation models.}
\label{tab:BLEU}
\end{table*}

\end{document}